\documentclass[10pt]{article}
\usepackage[letterpaper]{geometry}
\usepackage{hicss}
\usepackage{times}
\usepackage[none]{hyphenat}
\usepackage{url}
\usepackage{latexsym}
\usepackage{graphicx}
\graphicspath{{Figures/}}

\title{PPGN: Physics-Preserved Graph Networks for Real-Time Fault Location in Distribution Systems with Limited Observation and Labels}

\author{Wenting Li\thanks{This work is supported by the Center for Nonlinear Studies (CNLS) and a Laboratory Directed Research \& Development (LDRD) Exploratory Research (ER) grant at Los Alamos National Laboratory.}\\Los Alamos National Laboratory\\
 {\underline{wenting@lanl.gov}} \And
 Deepjyoti Deka \\Los Alamos National Laboratory\\
 {\underline{deepjyoti@lanl.gov}}\\ }

\date{}
\usepackage[frozencache=true,cachedir=minted-cache]{minted} 
\usepackage{amsmath}
\usepackage{graphicx}
\usepackage{array}
\usepackage{multirow}
\usepackage{epstopdf}
\usepackage{color}
\usepackage{booktabs}
\usepackage{float}
\usepackage{mathtools}

\def\V{{\mathcal{V}}}

\def\N{{\mathcal{N}}}
\def\G{{\mathcal{G}}}
\def\V{{\mathcal{V}}}
\def\P{{\mathcal{P}}}
\def\E{{\mathcal{E}}}
\def\S{{\mathcal{S}}}

\def\L{{\mathcal{L}}}
\def\C{{\mathbb{C}}}
\usepackage{amssymb}
\usepackage[numbers]{natbib}
\begin{document} 
\maketitle
\begin{abstract}
Electric faults may trigger blackouts or wildfires without timely monitoring and control strategy. Traditional solutions for locating faults in distribution systems are not real-time when network observability is low, while novel black-box machine learning methods are vulnerable to stochastic environments. We propose a novel Physics-Preserved Graph Network (PPGN) architecture to accurately locate faults at the node level with limited observability and labeled training data. PPGN has a unique two-stage graph neural network architecture. The first stage learns the graph embedding to represent the entire network using a few measured nodes. The second stage finds relations between the labeled and unlabeled data samples to further improve the location accuracy. We explain the benefits of the two-stage graph configuration through a random walk equivalence. We numerically validate the proposed method in the IEEE 123-node and 37-node test feeders, demonstrating the superior performance over three baseline classifiers when labeled training data is limited, and loads and topology are allowed to vary.
\end{abstract}

\subsubsection*{Keywords:}

Fault location, Graph neural networks, Limited observation, Low label rates, Distribution systems

\section{Introduction}
A modern power grid forms a critical infrastructure that delivers electricity for everyday energy consumption of society and the economy. In recent years, the expansion of random, intermittent distributed energy resources (DERs) such as wind and solar energy, particularly in low-voltage power grids, has increased the instability in the grid and resulted in surges, electric line failures, and other grid malfunctions \cite{NBHMTW09}.

However, localizing faults in power grids in real-time is faced with several practical challenges: low observability, unreliable estimates of system parameters, and the stochastic ambient environments due to random load variations and topology changes. The real-time observability in power grids has improved due to installing wide-area sensors at grid nodes, called phasor measurement units (PMUs) \cite{micropmu}, that collect time-synchronized measurements. This has motivated an interest in data-driven fault localization methods \cite{TWIA19, JYBFBT19, LLL19, LDCW19, CHZYH19} using PMU measurements.

These methods follow three research lines. Traveling-wave-based approaches are accurate and widely applied in the industry. The high-precision and synchronized measurements, however, require the measuring instruments to be installed everywhere, which hinders their extensive application \cite{TWIA19}. Another line of work relies on the physical property of data, such as the spatial relations of line impedance and the sparsity of fault currents, but these methods either require the full network observability \cite{LLL19} or high sampling rates (e.g., 10M Hz \cite{JYBFBT19}). The last research line is based on supervised machine learning to locate faults on the bus or line-level \cite{LDCW19, CHZYH19, AGR21}. These approaches show superior performance in efficiency and accuracy, especially in large-scale networks with low observability. Unfortunately, the insufficient availability of labeled data and the stochastic environment in practical power systems diminish the performance of such supervised methods.

Physics behind data has been incorporated into machine learning methods, so-called physics-informed machine learning (PIML), to enhance the interpretability and robustness to imperfect real data \cite{LW21, LD21}. However, such approaches that subtly combine physics with data-driven technology are lacking in the crucial problem of fault localization, for the realistic cases with limited labelled data.

To fill this gap, we analyze whether \textit{we can employ the unique physics of power grids to inform supervised data-driven fault localization algorithms and augment their robustness to challenges associated with realistic grid data.} In this paper, we are thus interested in fault localization in power grids, in the challenging but realistic regime of (a) sparse observations, and (b) system variability, with (c) low fraction of labeled training data. 

\textbf{Contributions:} We formulate a unique two-stage graph neural network architecture to locate faults in power grids with low observability and stochastic environments, using a small number of labeled data for training. Precisely, to address the issue of low observability, we inform $\G_{\text{I}}$ (GNN in the first stage) with the structure of the power grid by constructing an adjustable and novel adjacency matrix $A$. Meanwhile, in the second stage $\G_{\text{II}}$, we use a different adjacency matrix $B$ based on the statistical similarity of labeled and unlabeled datasets. This second GNN improves the localization accuracy when label rates are low. We theoretically interpret the functions of adjacency matrices $A$ and $B$ in stages I and II respectively, through equivalence with random walks. The proposed framework is validated in the IEEE 37 and 123-node test feeders \cite{K91}, in various scenarios through OpenDSS software \cite{DM11}. Our approach outperforms the existing algorithms by significant margins in accuracy and robustness to low label rates, load variations, and topology changes.

The remaining parts are organized as follows. Section \ref{sec:model} introduces the vital physics behind data and formulates the problem for the sake of some practical challenges. Sequentially, we present the two-stage graph learning framework in Section \ref{sec:method}. The following section demonstrates the benefits of the constructed adjacency matrices $A, B$. Section \ref{sec:simu} validates the effectiveness and advantages of our approach. Finally, conclusions and future works are discussed in Section \ref{sec:con}.

\section{Physics of Fault Currents and Problem Formulation} \label{sec:model}
\begin{figure}[!ht]
			\centering
			\includegraphics[width=0.4 \textwidth]{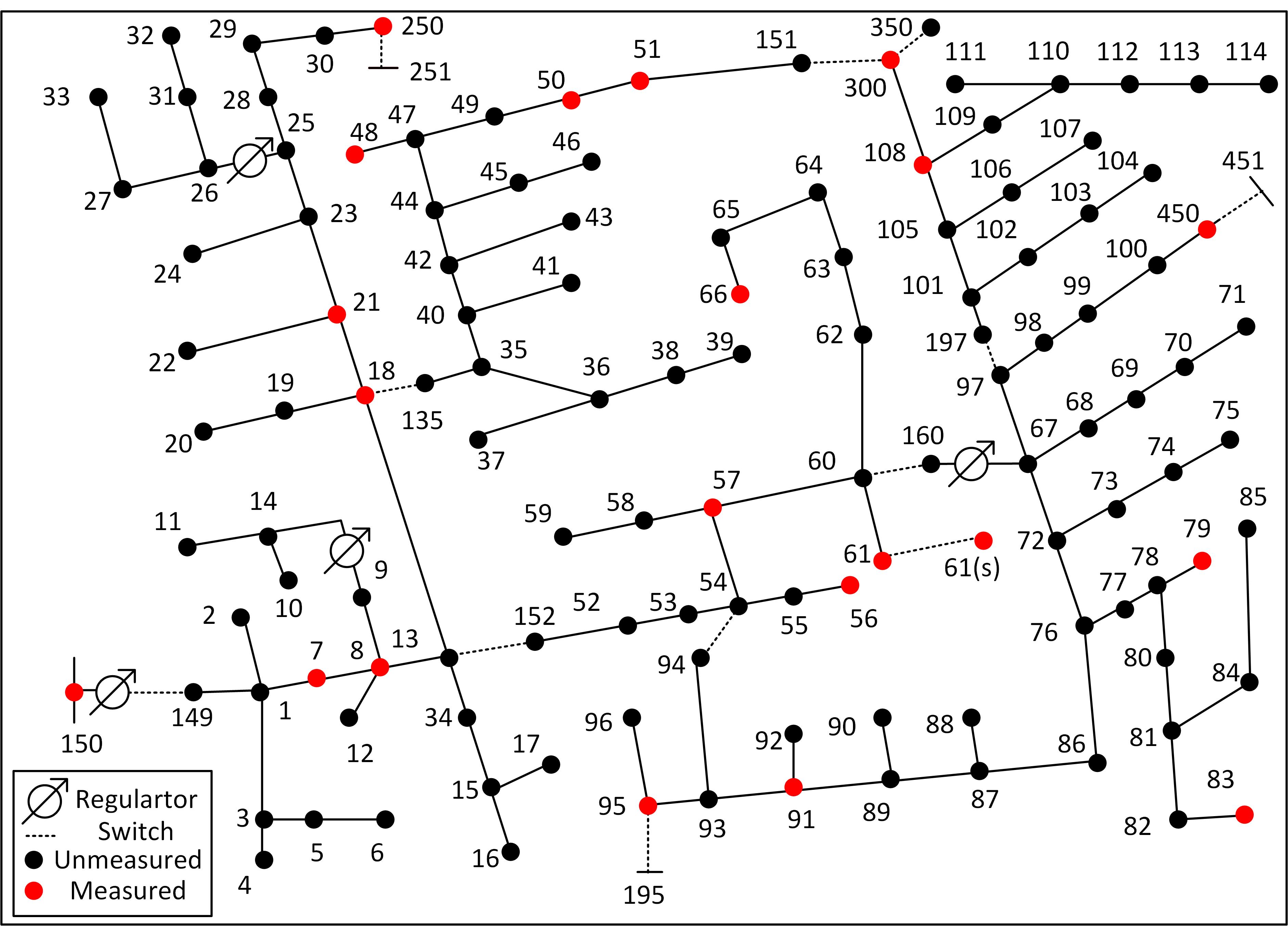}
			\caption{The IEEE 123-node test feeder}\label{123}
\end{figure}
Consider a power grid graph with $n$ nodes in the vertex set $\V$ and $l$ branches/edges in set $\E$, {for example, the IEEE 123-node test feeder in Figure.~\ref{123}.} Consider the three-phase voltages $u$ and currents $c$ at $s$ nodes in the set $\Omega \subset V$. Each data sample corresponds to a fault event, where the location of the fault is defined as its label. Interestingly, the nodal current (termed fault current) variations that arise due to a fault are sparse in nature, where the the nonzero values are closely related to the fault positions or labels \cite{ME18, JYBFBT19}. In this section, we will use this sparsity property to explain how the physical laws help reduce the label requirement.

\subsection{Sparsity of Fault Currents}\label{locality}
{When a fault occurs at $f$ (equivalent to a node) on the line between nodes $i$ and $j$ in the grid, we have voltages and currents at the faulted point and the nodes $k \neq f, k \in [1, n]$ denoted as $u_{f}^{abc}, c_{f}^{abc}, u_k^{abc}, c_k^{abc} \in \C^{3 \times 1} $ respectively. Let $Y_{ij} \in \C^{3 \times 3}$ be the admittance matrix between nodes $i$ and $j$ before the fault, while $Y'_{ij}$ is that during the fault. 	According to the Kirchhoff's law, we obtain the following equation \cite{LDCW19}, }
	\begin{align} \label{post1}
		&\underbrace{	\begin{bmatrix}
				Y_{11} & \cdots & \cdots & \cdots & Y_{1n}\\
				\cdots & \cdots & \cdots & \cdots & \cdots \\
				\cdots & Y_{ii} & \cdots & Y_{ij} \\
				\cdots & \cdots & \cdots & \cdots & \cdots \\
				\cdots & Y_{ji} & \cdots & Y_{jj} \\
				\cdots & \cdots & \cdots & \cdots & \cdots \\
				Y_{n1} & \cdots & \cdots & \cdots & Y_{nn} \\
		\end{bmatrix} }_{Y}
		\underbrace{	\begin{bmatrix}
				u^{abc}_{1} \\
				\cdots\\
				u_{i}^{abc} \\
				\cdots \\
				u_{j}^{abc} \\
				\cdots \\
				u^{abc}_{n}
		\end{bmatrix}}_U
		-
		\underbrace{	\begin{bmatrix}
				0\\
				\cdots\\
				\delta^{abc}_i \\
				\cdots \\
				\delta^{abc}_j\\
				\cdots \\
				0
		\end{bmatrix}}_{\Delta_{ij}}
		= \underbrace{	\begin{bmatrix}
				c^{abc}_{1} \\
				\cdots\\
				c_{i}^{abc} \\
				\cdots \\
				c_{j}^{abc} \\
				\cdots \\
				c^{abc}_{n}
		\end{bmatrix} }_C
		\nonumber \\
		&\Rightarrow Y U - \Delta_{ij} = C \\
		&\text{where~~~} \delta^{abc}_i = (Y_{ii} - Y'_{ii})u_{i}^{abc} + (Y_{ij} - Y'_{ij})u_{j}^{abc},\nonumber\\
		&~~~~~~~~~~~~~\delta^{abc}_j = (Y_{ji} - Y'_{ji})u_{i}^{abc} +(Y_{jj} - Y'_{jj})u_{j}^{abc}.\nonumber
		\end{align}
Notice that $\Delta_{ij} \in \mathbb{C}^{3n}$ is a \textit{sparse vector with the nonzero values corresponding to the two terminals $i,j$ of the faulted line}.

We also know that $U_0, C_0$, the voltages and currents on normal conditions, satisfy that $Y U_0 =C_0$. Let $\Delta U= U- U_0, \Delta C = C - C_0$ denote the changes in voltage and current respectively due to the fault. Using \eqref{post1}, we acquire the following relation between them:
	\begin{align}\label{feature}
		Y \Delta U & = \Delta C + \Delta _{ij}
	\end{align}
\textbf{Remarks:} The product of $Y $ and $\Delta U$, on the left side of \eqref{feature}, equals the linear combination of the node $k$'s neighbors weighted by the admittance, i.e., $ \Sigma_{j \in \N_k} Y_{kj} \Delta u_{j}^{abc}, k=1, \cdots, n$, where $\N_k$ denotes the set of nodes connected with $k$. On the right side of \eqref{feature}, $\Delta_{ij}$ only has nonzero values at the nodes $i, j$ connected with the fault point, while $\Delta C$ is trivial since loads have a small chance to change dramatically during the fault \cite{MAE15}. When all buses are known, the weighted voltage variations $ \Sigma_{j \in \N_k} Y_{kj} \Delta u_{j}^{abc} $ are significant if $k$ is near the fault. When the observability is low, the measured buses partially include the information of fault location, and a learning strategy benefits by extracting the relations between the partially weighted voltages and the location of faults. Therefore, we formulate the fault location as a learning problem as follows.

\subsection{Problem Formulation}
We consider a $N$ length data-set of voltage magnitudes and voltage angles in $a,b,c$ three phases from $s < n$ measured nodes in the grid, i.e., $X^p \in R^{n \times 6} = [ V^a, \theta^a, V^b, \theta^b, V^c, \theta^c], p = 1, \cdots, N$, only the $s$ entries of $X^p$ corresponding to the measured nodes have nonzero values. Here unmeasured nodes are given a value $0$. Additionally, we have partial labels denoting the location of the faults, $y^p \in \{1, \cdots, c\}$ for some datasets $p = 1, \cdots, m$, where $m \ll N$. We target at efficiently predicting the location of unknown faults regardless of fault types and fault impedance. Note that $m\ll N$ implies that only a few number of datasets are labeled with the true fault location while others are not.

Fault location is equivalent to a classification problem \cite{LDCW19}, but practically this classifier faces more challenges. Though Section \ref{locality} demonstrates the effect of power system topology in estimating faults via $\Delta U$, the sparse observation of power network and low label rates hinders us from directly applying a conventional graph neural network (GNN) \cite{SGAHM08} classifier on the the power grid topology. Furthermore, the non-static measurements of power grids demand the classifier to be robust to out of distribution (OOD) data \cite{CCTS20}, due to changing load dispatch, random fault impedance, and topology changes.

	\section{Proposed Graph Learning framework} \label{sec:method}
	\begin{figure}[thb]
		\centering
		\includegraphics[width=0.46\textwidth]{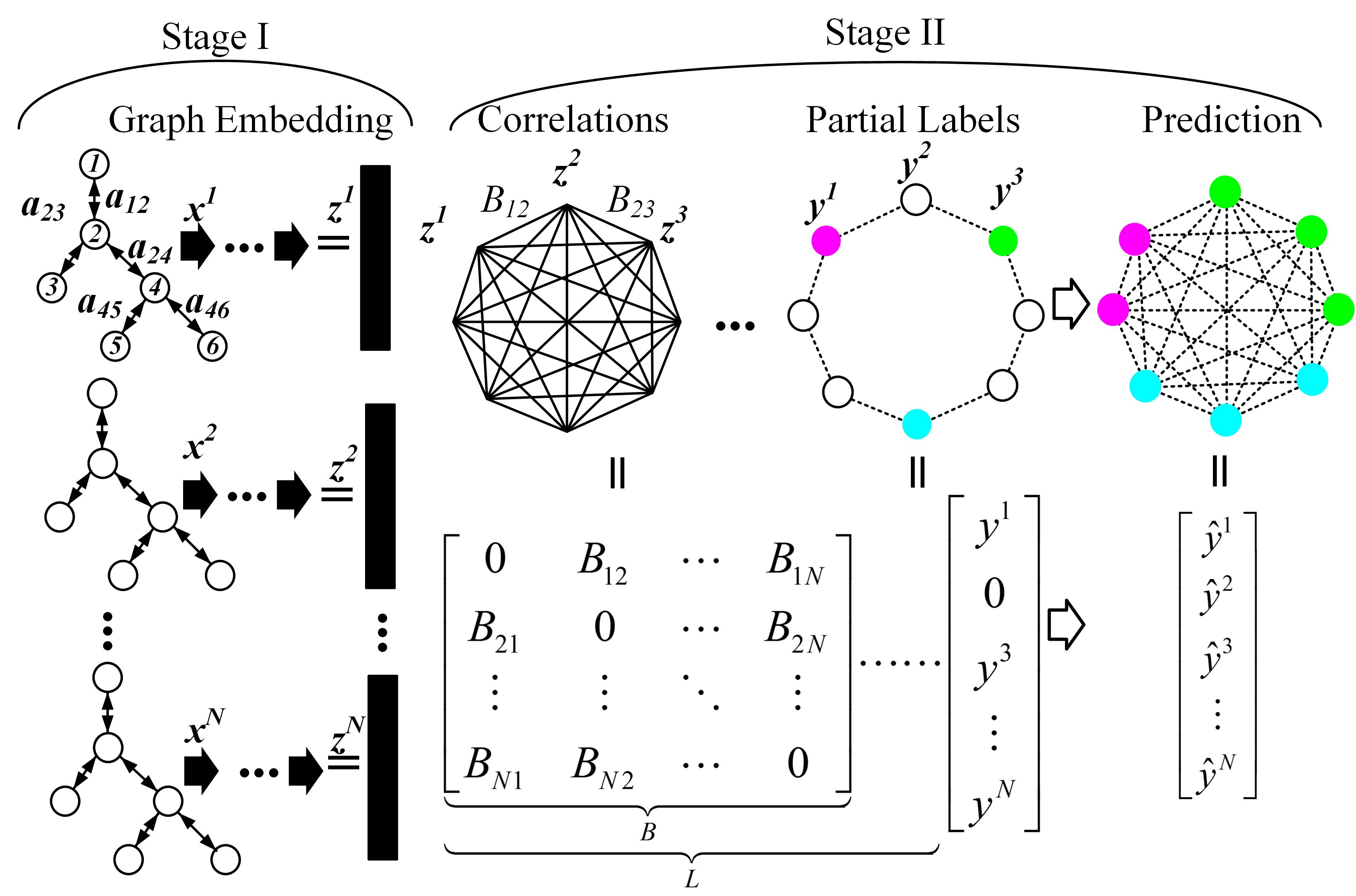}
		\caption{The structure of PPGN} \label{structure1}
	\end{figure}
Figure~\ref{structure1} shows our two-stage Physics-Preserved Graph Network (PPGN) learning framework for fault location. {In Graph Stage I, named as $\G_{\text{I}}$, we first learn a graph embedding to represent both the node features and topology structure of the power grid by locally aggregating the observed nodes, and then predict the faulty node through the global transformation. The core distinction of  $\G_{\text{I}}$ is the \textit{adjustable adjacency matrix} for local aggregation. Such adjacency matrix, different from the conventional GNN \cite{KW16, HWZJ17}, handles the challenges of sparse observability and enlarges the capability of location prediction.} Following Stage I, we design the Stage II with another graph neural network $\G_{\text{II}}$ using correlations between the available labeled and unlabeled datasets, to further improve the location accuracy. The novel adjacency matrix in $\G_{\text{II}}$ is based on the learnt graph embedding from Stage I. The detailed theoretical analysis of these two adjacency matrices for $\G_{\text{I}} $ and $\G_{\text{II}}$ is presented in Section \ref{sec:analysis}.
\subsection{Stage I: Graph Embedding Learning}\label{stageI}
When the measured nodes of the power grid are sparse, it is difficult to learn informative embedding of the node without measured neighbors. Instead, our key idea is to construct an adjacency matrix $A \in R^{n \times n}$ based on {Dijkstra's shortest path} \cite{DO59} between any pair of nodes to ensure that each node is observable. Then we accomplish graph embedding with two major procedures: local aggregation and global transformation.\\
\noindent \textbf{3.1.1 Local Aggregation}
$K$ hidden layers map the $p$th data matrix $X^p $ into hidden variables $H^k \in R^{n \times n_k} =[h_1^k, \cdots, h_n^k], k=1, \cdots, K$ with the shared weight matrix $W^k \in R^{2n_{k-1} \times n_{k}}$ among all nodes, and $H^0 = X^p$. The $i$th row and $j$th column of A represents the correlation between nodes $i, j$ and is defined as
\begin{align}\label{adj}
		a_{ij} & = \begin{cases}
			\exp(-\frac{d^2_{i,j}}{\delta_i^2})	 & \text{if }j \in \N_i^{k_{\text{I}}}\\
			0 & \text{else}
		\end{cases}
\end{align}
where $d_{ij}$ is the shortest path between nodes $i, j$, $\delta_i = \frac{1}{k_{\text{I}}} \Sigma_{j \in \N_i^{k_{\text{I}}}} d_{ij}$, and $ \N_i^{k_{\text{I}}}$ consists of the $k_{\text{I}}$ nearest neighbors of node $i$, {where $k_{\text{I}}$ is chosen to ensure that each unobserved node has at least one observed neighbor.} We symmetries the matrix $A$ by $a_{ij} = \max(a_{ij}, a_{ji} )$. 

The update rule of the $k$th layer is:
\begin{align}\label{local}
		h_i^k & = \sigma(\{h_i^{k-1} || \text{Aggregate}_{j \in \N_i} (h_j^{k-1} \tilde{a}_{ij}) \} W^k)
	\end{align}
where $ ||$ means to concatenate the two vectors, $ \text{Aggregate}_{j \in \N_i}(h_j^{k-1} \tilde{a}_{ij}) = \frac{1}{|\N_i|}\Sigma_{j \in \N_i}(h_j^{k-1} \tilde{a}_{ij})$, $\sigma(x) = \max(0,x)$, $\tilde{a}_{ij}$ is the normalized $a_{ij}$ such that $\Sigma_{j \in \N_i} \tilde{a}_{ij} = 1$, and the $\N_i$ is the neighborhood of node $i$. \\
\noindent \textbf{3.1.2 Global Transformation }
	Global transformation converts the hidden variables of all the local nodes into prediction probability of the whole data sample as the graph embedding.
	Firstly, the vectorized hidden variables $\hat{h}^K \in R^{n n_K}$ go through fully connected layers with the trainable weights $W^f \in R^{nn_K\times 2n}, b^f \in R^{2n}, W^o \in R^{2n \times n}, b^o \in R^n$ to become the vector $f \in R^{n}$. Then the output layer transforms $f$ into graph embedding $z^p \in [0,1]^n$ with $W^o \in R^{2n \times n}, b^o \in R^n$ for the $p$th data sample as follows,
	\begin{align}\label{global}
		z^p_i & =
		\frac{\exp (f_i )}{\Sigma_{j =1}^n\exp (f_j )}, \quad	f =( \hat{h}^K W^f + b^f ) W^o+ b^o
	\end{align}
	where $z^p_i, f_i$ are the $i$th entry of $z^p$ and $f$ respectively. \\
\noindent \textbf{3.1.3 Loss Function of Stage I}
	\begin{align}\label{obj_I}
		\L(\Theta_{\text{I}} ) & = - \Sigma_{p = 1}^m y^p \log (z^p) + \lambda_{\text{I}} \lVert \Theta_{\text{I}} \rVert
	\end{align}
	The first term of \eqref{obj_I} is the cross entropy of $y^p$ and $z^p$ for available labeled data samples, and the second is the regularization term to augment the generalization capability, where $ \lVert \Theta_{\text{I}} \rVert $ is the $l_2$-norm of all the trainable parameters $\Theta^I =\{W^1, \cdots, W^K, W^f, b^f, W^o, b^o\}$ of $\G_{\text{I}}$ with the hyper-parameter $\lambda_{\text{I}}$. By minimizing \eqref{obj_I}, the $\Theta_{\text{I}}$ is automatically learned by back-propagation. \\
\noindent \textbf{3.1.4 Practical Training Technique}\label{T1_T2}
{Inspired by Yang et.al. \cite{YCS16}, we alternatively train $\G_{\text{I}}$ through the local aggregation and global transformation procedures for $T_1$ and $T_2$ epochs respectively. Empirically, this alternative training speeds up the convergence and accomplishes higher classification accuracy. The detailed discussion is in Section \ref{G}.
}
	\subsection{Stage II: Label Propagation}\label{z}
	To further increase the location accuracy when given a number of unlabeled datasets, we build the correlation matrix $B \in R^{N \times N}$ of the labeled and unlabeled datasets, as the adjacency matrix of the graph $\G_{\text{II}}$. Note that each vertex of $\G_{\text{II}}$ represents one data sample. The intuition is that faults that occur at the same or nearby locations share similar physical characteristics, as the analysis in Section \ref{locality}, resulting in similarity of datasets. {Using the learned graph embedding $z^p$ in $\G_{\text{I}}$ of various data samples, we establish the correlation matrix $B$. Then our graph model $\G_{\text{II}}$ propagates labels to the unlabeled data samples through graph convolutional layers (GCL) \cite{KW16}.} \\
	\noindent \textbf{Adjacency Matrix $B$:} The critical purpose of $B$ is to learn useful correlations among data samples while cutting off misleading correlations. For that we use $z^p$, the output of Stage I. Precisely, we first zero out the entries of $z^p$ that correspond to nodes far beyond the Stage I predicted location $p^*= \arg \max_{i} z^p_i$, and obtain vector $\hat{z}^p$. Let $\S_{p^*}$ be the set of nodes physically connected with $p^*$ in the original power grid. Then
	\begin{equation}\label{z_hat}
		\hat{z}^p_k = \begin{cases} 0 &\text{if }k \notin \S_{p^*}\\
			z^p_k &\text{otherwise}
		\end{cases}
	\end{equation}
	Second, we calculate the similarity $s(p,q)$ of embedding between any pair of data samples $\hat{z}^p, \hat{z}^q $ through distance metrics. Here we apply the subspace angle $s(p,q) = \frac{(\hat{z}^p, \hat{z}^q)}{\lVert \hat{z}^p \rVert_2 \lVert \hat{z}^q \rVert_2 }$ \cite{SEC14} as distance metric. The entry at the $p${th} row and $q${th} column of $B$ is then defined as
	\begin{equation}\label{B1}
		B_{pq} = \begin{cases}
			s(p,q) & \text{if $q, p$ are similar to each other} \\
			0 & \text{else}
		\end{cases}
	\end{equation}
	where ``$q, p$ are similar to each other '' means that the $s(p,q )$ is among the largest $k_{\text{II}}$ values of $\{s(p, q'), q' \in [1, N]\}$ or $\{s(p', q), p' \in [1, N]\}$. Significantly, $B$ becomes a sparse matrix with non-zero entries restricted to data-points at close graphical locations. This helps accelerate the training process when $k_{\text{II}} \ll N$. We give a theoretical explanation for $B$ in Section \ref{B}. Once $B$ is estimated, we use it in a Graph Convolutional Network.\\
	\noindent \textbf{3.2.1 Graph Convolutional Layers }
	Reshape all the raw data samples $X^p, p=1,\cdots, N$ to form matrix $C^0 \in R^{N \times 6n}$ as the input of the Graph Convolutional Layers (GCL).
	\begin{equation}
		C^l = \sigma(D^{-\frac{1}{2}}	\hat{B} D^{-\frac{1}{2}}C^{l-1} W_{\text{II}}^{l})
	\end{equation}
	where $C^l, l = 1, \cdots, L$ is the $l$th output of the hidden layer with weights $W_{\text{II}}^{l}\in R^{n_{l-1} \times n_l}$, $\hat{B} = I_N + B $ and $D$ is the degree matrix corresponding to $\hat{B}$ for normalization. \\
	\noindent \textbf{3.2.2 Output Layer}
	The output layer performs the linear regression on the $c^p$, the $p$th row of $C^L$, with the weight $W_{\text{II}}^o \in R^{n_{L} \times n }$ and the bias $b_{\text{II}}^o \in R^{n}$ to obtain $g^p \in R^n$, and then converts $g^p$ to be the prediction probability $ \hat{y}^p \in R^n$ through a softmax function as follows, 
	\begin{align}
		\hat{y}^p_i & = 	\frac{\exp (g^p_i)}{\Sigma_{j=1}^n \exp (g^p_j )}, \quad g^p = c^p W_{\text{II}}^o + b_{\text{II}}^o
	\end{align}
	where $\hat{y}_i^p, g^p_i$ are the $i$th entry of $\hat{y}^p$ and $g^p$ respectively. \\
	\noindent \textbf{3.2.3 Loss function of Stage II} We use the regularized cross entropy loss function as
	\begin{align}\label{obj_stageII}
		\L'(\Theta_{\text{II}}) & = -\Sigma_{p=1}^m y^p \log (\hat{y}^p) + \lambda_{\text{II}} \lVert \Theta_{\text{II}} \rVert
	\end{align}
	where $\Theta_{\text{II}} = \{W_{\text{II}}^1, \cdots, W_{\text{II}}^L, W_{\text{II}}^o, b^o \}$ includes all the trainable parameters of $\G_{\text{II}}$.
	\section{Theoretical Interpretations}\label{sec:analysis}
We interpret the graph learning in the two stages through the random walk equivalence \cite{WL20, XLTSKJ18}. Here \textit{a walk} steps from one node randomly into its neighbor defined by the adjacency matrix of the graph. From the point of this view, we demonstrate the advantages of our constructed $A$ and $B$ in augmenting visibility and location accuracy.
	\subsection{Construct $A$ to Improve Visibility} \label{walk}
	\begin{figure}[thb]
		\centering
		\includegraphics[width=0.12\textwidth]{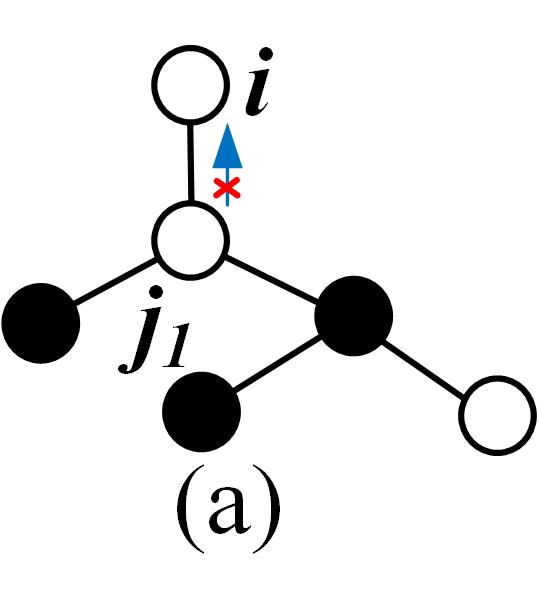}
		\includegraphics[width=0.12\textwidth]{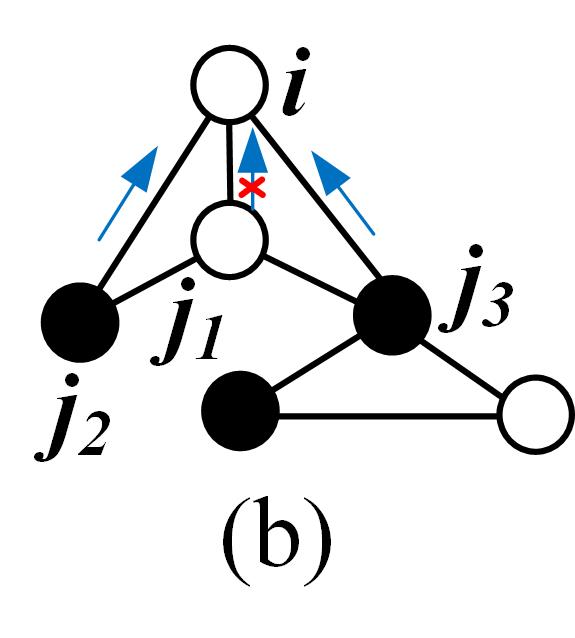}
		\caption{Example: Filled nodes are measured, and void nodes are unmeasured. When $k_{\text{I}} = 2, K=1$, node $i$ only has one connected node $j_1$ in (a), but $i$ has three neighbors $j_k,k=1,2,3$ in (b) defined by our $A$, which ensures that there are some paths from the unmeasured nodes to the measured nodes.}\label{path}
	\end{figure}
The construction of $A$ determines the learning path in the stage I. We rigorously derive the information flow along the path through node influence.\\
	\noindent \textbf{Node Influence of $\G_{\text{I}}$:}
	The \textit{node influence}, denoted as $I(h^k_i, h^0_j; k )$, measures the influences of the node $j$'s input, $h^0_j$, on node $i$'s learned hidden variable, $h^k_i$ after $k$ layers \cite{XLTSKJ18}, which quantifies the variations of $h^k_i$ \textit{when $h_j^0 $ changes, }
	\begin{align}\label{S}
		I(h^k_i, h^0_j ; k) & = \partial{h_i^k} / \partial{h_j^0} \nonumber \\
		& = \text{Pr}(\text{Walk from $i$ to $j$ with $k$ steps}) \nonumber \\
		& = \Sigma_{\P \in \P^{i \rightarrow j}_{k}} \prod_{e \in \P} \tilde{a}_e
	\end{align}
	where {Pr} is the abbreviation of probability, $\P_{k}^{i \rightarrow j}$ represents the path from $i$ to $j$ with $k$ steps, $a_e$ is the weight of the edge $e$ defined by $A$ that $\tilde{a}_e = \tilde{a}_{i'j'}$ when $e = (i',j')$. As the learning process of each layer is independent, the probability of choosing one path $\P$ is the product of $\tilde{a}_e, e \in \P$.

	The information collected by each node via $\G_{\text{I}}$ is implied by the \textit{total node influence }$I(h_i^K)$ of the node $i$ after $K$ layers,
	\begin{align}
		I(h_i^K) & = \Sigma_{j \in \N^K_i, j \in \Omega} 	I(h^K_i, h^0_j; K ) \nonumber \\
		& = \Sigma_{j \in \N^K_i, j \in \Omega} 	 \Sigma_{\P \in \P^{i \rightarrow j}_{K}} \prod_{e \in \P} \tilde{a}_e
	\end{align}
	where $\N_i^K$ denotes the $K$-hop neighborhood of node $i$, $\Omega$ is the set of measured nodes. Therefore, we conclude that the information obtained at node $i$ is richer if more paths from the observed nodes to node $i$.
	Following this principle, we construct $A$ to ensure that each unobserved node has some path from the nearby observed nodes even though its immediate neighbors are unmeasured. Figure~\ref{path} (a) shows one simple example to illustrate the distinctive learning paths when using the physical topology in (a) and that using our $A$ in (b).
	\subsection{Construct $B$ to Enhance the Exact Prediction Probability} \label{B}
 $B$ characterizes the correlations of labeled and unlabeled data samples to improve the prediction probability. We first show the random-walk interpretation of the node influence in $\G_{\text{II}}$, and then introduce how we increase correct prediction probability by cutting off misleading correlations. The effects of $B$ are also validated by experiments in Section \ref{label}. \\
		\noindent \textbf{Node Influence of $\G_{\text{II}}$:}
		In $\G_{\text{II}}$, the $p$th node corresponds to the data sample $X^p$, and we care about the influence of the known labeled data samples on the unlabeled ones. The node influence of the labeled data sample $q $ on the unlabeled data sample $p $ after $l$ layers is $	I'(C^l_p, C^0_q; l)$.
		\begin{align}\label{S'}
			I'(C^l_p, C^0_q; l) & = \partial{C^l_p} / \partial{C^0_q} \nonumber \\
			& = \text{Pr (Walk from $p$ to $q$ with $l$ steps) }\nonumber \\
			& = \Sigma_{\P \in \P^{p\rightarrow q}_{l}} \prod_{e \in \P} \bar{B}_{e}
		\end{align}
		where $\bar{B}_e = \bar{B}_{p'q'}$, $e = (p',q')$ is the edge $e$ between $p'$ and $q'$, and $\bar{B}_{p'q'} $ is the normalized weight that $\Sigma_{q' = 1}^N \bar{B}_{p'q'} = 1$.

	The expectation of the total influence of node $q$ on $p$ after $L$ hidden layers is defined as (by Theory 1 in \cite{XLTSKJ18},
		$$I'_p(q; L)= {\Sigma_{q }	I'(C^L_p, C^0_q; L) }/{\Sigma_{r}		I'(C^L_p, C^0_r; L) }$$
		is equivalent to the probability that data sample $p$ has the same label with $q$ after $L$-step random walk, i.e.,
		\begin{align}\label{prob}
			\text{Pr}(\hat{y}^p= y^p) & = E(\frac{\Sigma_{q: y^q= y^p}	I'(C^L_p, C^0_q; L ) }{\Sigma_{r: y^r \in [1,c]} 	I'( C^L_p, C^0_r; L)}) \\
			& = E(\frac{\Sigma_{q: y^q= y^p} \Sigma_{\P \in \P^{p\rightarrow q}_{L}} \prod_{e \in \P} \bar{B}_{e} }{\Sigma_{r: y^r \in [1,c]} \Sigma_{\P \in \P^{p\rightarrow r}_{L}} \prod_{e \in \P} \bar{B}_{e}})
		\end{align}
		where $q: y^q= y^p$ denotes those data samples $q$ have the same labels with $y^p$, and $r: y^r \in [1,c]$ represents any data sample $r$.
		Hence, the correct prediction of the unlabeled data sample $p$ has high probability if (1) The paths from the data samples with the same labels $y^p$ are more than those with other labels; (2) the number of labeled data samples with $y^p$ is significant. \\

		\noindent \textbf{Cut off the Unrelated Paths through $B$: }
		We construct $B$ via $\hat{z}^p, p=1, \cdots, N$ rather than the raw datasets $X^p$ to produce more zero entries that cut off the paths from mismatching labels, resulting in an increase in the accurate prediction probability. If we calculate $B$ with $X^p$, the correct prediction probability $\text{Pr}(\hat{y}^p= y^p)$ is given by \eqref{prob}. However,
		if we use $\hat{z}^p$ as input, it is clear from \eqref{z_hat} and \eqref{B1} that $s(p, q)$ for data-sets $p,q$ is nonzero only if the prediction $q^*$ and $p^*$ are within two-hops of each other. Thus, the physical distance between the true labels of data-set $p$ and $q$ should be within four hops if $s(p,q)$ is not zero. The correct prediction probability, in that case, is thus larger as the $p$th data sample only has the access to partial data samples $r$ whose true labels $y^r \in \N^{{4}L}_{y^p}$, i.e.,
		\begin{align}
			P'(\hat{y}^p = y^p) & = E(\frac{\Sigma_{q: y^q = y^p} I'(X^p, X^q ; L)}{\Sigma_{r: y^r \in \N^{4L}_{y^p}}I'(X^p, X^r; L)}) \\
			& \geq 	\text{Pr}(\hat{y}^p= y^p) \nonumber
		\end{align}
		As we also control the total number of nonzero values of each row of $B$ to be no more than $2k_{\text{II}}$, thus $|\N^{4L}_{y^p} | \leq 2k_{\text{II}} < |\V|$, where $| \cdot |$ denotes the size of a set. Therefore, we improve the correct prediction probability by constructing $B$ using the learned graph embedding.
\section{Numerical Experiments}\label{sec:simu}
We implement the proposed framework in the 123-node test feeder \cite{K91}, simulated by OpenDSS \cite{DM11}. This test feeder is typically composed of grid components such as voltage regulators, overhead/underground lines, switch shunts, and unbalancing loads. Our approach shows high performance for various types of faults at low label rates, outperforming three well-known baselines by significant margins. Moreover, we validate the robustness to topology changes and load variations, and analyze the effects of different stages. In addition, our graph learning framework can easily adapt to another system, the IEEE 37-node test feeder, where we demonstrate the superior location performance using the proposed training strategy in Section \ref{T1_T2}. The codes and datasets are available at \url{https://github.com/Wendy0601/PPGN-Physics-Preserved-Graph-Networks}.
\subsection{Implementation Details }
		The graphical structure our testing system is shown in Figure~\ref{123}, where 21 out of the 128 nodes are measured marked as red. Nine pairs of nodes are connected by switches or regulators at the same locations. Thus a total of 119 possible fault positions exist, which are represented by the labels $y^p \in \{1,\cdots, c\}, c= 119$. We simulate $N = 24480$ data samples including single phase to ground (SPG) faults, phase to phase (PP) faults, and Double-phase to ground (DPG) faults at all the three phase nodes with fault impedance varying from $0.05\Omega \sim 20 \Omega$. In the industrial practice, training data samples can be acquired either from historical datasets or by some advanced data generation techniques, such as \cite{CWKZ18}.

		We implement three best baseline classifiers: fully connected neural networks (NN), convolutional neural networks (CNN), and graph convolutional neural networks (GCN). Our NN has two rectified linear unit (ReLU) layers, each of which reduces the dimension of the input by one half; CNN has four ReLU convolution layers with filters of size $2 \times 2$ and depth of $8, 8, 16,16$. Each convolution layer is followed by batch normalization and maximum pooling layers; GCN has three convolutional graph layers with filters of size $32$. The baseline classifiers all apply the cross entropy loss function with $l_2$ norm regularization. We apply Adam, a stochastic gradient descent based optimizer \cite{KBJ14}, with learning rate being 0.001 to train the classifiers.

		The loads for each data sample are random, following a typical load shape, and the expectation of load variations at one node is 0.53 per unit (p.u.). Each data sample is a matrix $X^p \in R^{128 \times 6}$, where only measured nodes have nonzero values. To validate the robustness to OOD data, we simulate another nine sets of data samples covering various system topology or load variations as shown in Section \ref{robust}, and each set includes 12240 faults of all types and locations. We normalize each data sample $X^p$ by subtracting the mean values and scaling with the standard deviation of all datasets. \\
\noindent \textbf{Structures of $\G_{\text{I}}, \G_{\text{II}}$}\label{structure}
		$\G_{\text{I}}$ has three hidden layers with $W^k \in R^{32 \times 32}, n_k = 32, k=1, \cdots, K = 3$. We set $k_{\text{I}} = 3 \ll n$ considering the sparse infrastructure of power grids, to construct matrix $A$ (see Eq.~\ref{adj}). The hyper-parameters $\lambda_{\text{I}} = 5 \times 10^{-3}$, $ T_1 = 10, T_2 = 10$, and the learning rate is 0.001. $\G_{\text{II}}$ has two layers with $n_l = 3n$ for $l=1,2$. We take hyper-parameter $k_{\text{II}} = 120$ for $B$ (see Eq.~\ref{B1}). The learning rate for $\G_{\text{II}}$ is 0.001 and $\lambda_{\text{II}} = 5 \times 10^{-5}$. We train the proposed model with Adam and implement the the structure through Pytorch \cite{PGM19}. Based on a MacBook Pro with CPU of 2.4 GHz 8-core Intel i9 series, memory of 32 GB, the per-iteration running time of the proposed graph neural network $\G_I$ is 0.03 seconds, which becomes 0.02 seconds if using 1 NVIDIA Tesla GPU. \\
\noindent \textbf{Performance Metrics }
		We adapt three performance metrics: F1-score, location accuracy rate (LAR), and {LAR}$^{1-\text{hop}}$ \cite{PVGM11}. The definitions of these three metrics are based on four basic concepts: True Positive (TP$_i$) is the number of correctly predicted samples of location $i$; False Positive (FP$_i$) is the number of wrongly predicted samples of location $i$; True Negative (TN$_i$) is the number of correctly predicted samples of locations rather than $i$; False Negative (FN$_i$) is the number of wrongly predicted samples of locations rather than $i$. Let $\text{TP}'_i$ denote the number of data samples where the predicted node is in the immediate neighborhood of the true fault node $i$. Then \textit{Precision} is $P_{i} = {\text{TP}_i}/{(\text{TP}_i+\text{FP}_i)}$, \textit{Recall} is $R_i = {\text{TP}_i}/( {\text{TP}_i+\text{FN}_i} )$, \textit{F1-score }is ${F}_i = 2 {R_i P_i }/{(R_i + P_i)}$, $ \text{LAR}_i =\text{TP}_i/(\text{TP}_i+\text{FP}_i + \text{TN}_i+\text{FN}_i)$, and $ \text{LAR}_i^{1-\text{hop}} =\text{TP}'_i/(\text{TP}_i+\text{FP}_i + \text{TN}_i+\text{FN}_i)$.
		The notations LAR and LAR$^{1-\text{hop}}$ are the average of LAR$_i$ and LAR$^{1-\text{hop}}_i, i \in [1,c]$ of all the locations.

		\subsection{Performance Comparison}
		\begin{figure}[thb]
			\centering
			\includegraphics[width=0.23 \textwidth]{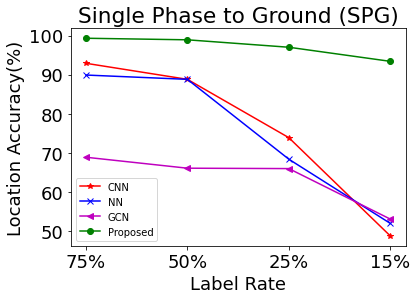}
			\includegraphics[width=0.23 \textwidth]{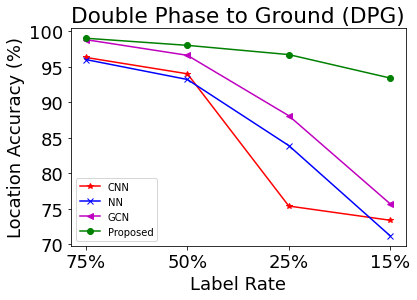}
			\includegraphics[width=0.23 \textwidth]{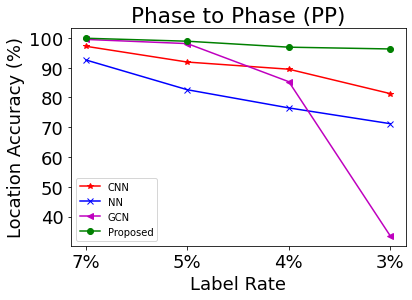}
			\includegraphics[width=0.23 \textwidth]{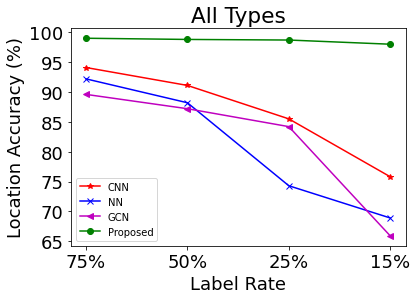}
			\caption{\small{LAR at Different Label Rates }} \label{fig:spg}
		\end{figure}
		We compare our model with three baseline classifiers (CNN, NN, GCN)
		in Figure~\ref{fig:spg}, which includes the location performance for different types of faults respectively. Because PP faults are less impacted by the changes of grounded impedance, we find that locating the PP faults is much easier than others. We show the location performance for PP faults when label rates are from 7\% to 3\% and 75\% to 15\% for others. Note that the power grid operator can first determine the type of each fault by other approaches, such as \cite{LW19, LWC18}.

		We find that when the label rates are high, {GCN and CNN can achieve comparable performance with the proposed model, but the high performance depends on the sufficient training data, and their LARs decline dramatically when label rates become low. }On the contrary, the proposed algorithm shows more stable and accurate performance, outperforming the baselines by significant margins at realistic low label rates.

		\subsection{Two-stage Location Performance}
		\begin{table}[!ht]
			\centering
			\caption{Location Performance of All Types} \label{cmp_all}
			\begin{tabular}{c c c c c }
				\hline \hline Label Rate $\beta$& 75\% & 50\% & 25\% & 15\% \\
				\hline F1 Score (\%) & 98.7 & 98.5 & 	98.4	 & 97.6 \\
				\hline LAR (\%) & 99.0 & 	98.8& 	98.7 & 	98.0	\\
				\hline $	\text{LAR}^{1-\text{hop}} $ (\%)& 100.0 & 	100.0 & 	99.96 & 	99.90 \\
				\hline
			\end{tabular}
		\end{table}
		We train $\G_{\text{I}}$ at different label rates $\beta$, which is the ratio of the number of training data samples to $N$. Table \ref{cmp_all} reports the location performance in terms of the three metrics when $\beta$ varies from 75\% to 15\%.
		F1 Scores and LARs of our model are stable and remain higher than 97\% even only $\beta=$15\%. Note that the labeled data samples for training are randomly selected for each location to avoid the issue of data imbalance \cite{THKG20}. Crucially, the LAR$^{1-\text{hop}}$ is close to 100\%, indicating that the predicted node is in the 1-hop neighborhood of the fault node with a high probability.
		\subsection{Performance at Different Stages When Label Rates are Low}\label{label}
		\begin{table}[!ht]
			\centering
			\caption{Location Performance of Different Stages}\label{II}
			\begin{tabular}{c c c c }
				\hline \hline Types of Faults & SPG & DPG & PP \\
				\hline Label Rate & 15\% & 	15\% & 	3\% \\
				\hline Stage I Only (LAR)& 92.9	& 92.7&	95.7 \\
				\hline 	 Stage II Only (LAR) & 30.7 & 39.6 & 78.2 \\
				\hline Stage I + II (LAR)& \bf 93.3& \bf	93.5& \bf	96.3 \\
				\hline Stage I Only ($	\text{LAR}^{1-\text{hop}} )$ & 94.8	&95.1&	96.6 \\
				\hline 	 Stage II Only ($	\text{LAR}^{1-\text{hop}} )$ & 37.3 & 	46.3 & 82.3\\
				\hline Stage I + II ($	\text{LAR}^{1-\text{hop}} )$& \bf 99.7& \bf	98.9& \bf	99.8 \\
				\hline
			\end{tabular}
		\end{table}
		Table \ref{II} shows the performances at different stages when label rates are low, where ``Stage I Only'' denotes that we only train $\G_{\text{I}}$ to locate faults; ``Stage II Only'' means that we directly utilize the input data $X^p$ rather than the embedding $z^p$ to construct the matrix $B$ and then locate the unknown faults. Note that ``Stage II Only'' is the typical manner of label propogation strategy for the semi-supervised learning in the computer vision domain \cite{HWZJ17}. ``Stage I + II'' represents our proposed graph learning framework.

		Comparing with the ``Stage I Only'' and ``Stage II Only'', we observe the superior performance of combining stage I and II. Also, the low LAR of ``Stage II Only'' indicates the significance of constructing $B$ using $z^p$ rather than $X^p$. The high accuracy of ``Stage I Only'' ensures the learned $z^p$ is reliable. Therefore, combination of the stages I and II benefits the location accuracy further.

		\subsection{Robustness to Out of Distribution Data}\label{robust}
		We validate the robustness of different classifiers to OOD data due to load variations and topology changes. Note that the performance of each classier here is for the best model in Figure~\ref{fig:spg}, \textit{but without retraining}.

		Table \ref{load} demonstrates the robustness to load variations, where $\Delta p$ denotes the averaged load variation per unit (p.u.) at each node with a load. The $\Delta p$ for training data is 0.53, and here we increase it up to 0.74 p.u., which immediately causes the measurements to change to different extent. Compared with other classifiers, our proposed method achieves the highest accuracy with less variations and hence is more robust to the load variations.

		Also, we change the topology by varying the states of eight switches. 
		Under normal conditions, the first six switches are closed, and the other two are open.
		Tables \ref{topology} reveals the testing accuracy of the datasets with various switch states, where ``Close 7\&8'' denotes that we close the switches 7 and 8 from open states, and ``Open 1-6'' means that we open the first 6 closed switches. Note that the challenge here is not only the change of topology, but also the data variations caused by topology changes. Still, the proposed one demonstrates better accuracy in all scenarios.
		\begin{table}[!ht]
			\centering
			\caption{LAR of SPG, DPG, PP When All the Loads Vary in Different Ranges} \label{load}
			\begin{tabular}{c c c c c c c }
				\hline \hline SPG & \begin{tabular}{@{}cccccc@{}}
					$\Delta p$ (p.u.) & 0.53 & 0.58 & 0.64 & 0.69 & 0.74\\
					\hline 	CNN& 93.9 & 	85.3 & 	84	 & 83.9	 & 82 \\
					\hline NN & 92.5 & 	80 & 	77.4 & 	76.7 & 	74 \\
					\hline GCN & 64.3 & 	57.7 & 	56.4 & 	55.6 & 	55.1 \\
					\hline Proposed & \bf 98.9 & \bf 	96.6	 & \bf 	96.3 & \bf 	95.8& \bf	95.1
				\end{tabular} \\
				\hline
				DPG & \begin{tabular}{@{}cccccc@{}}
					CNN& 96.5 & 	88.3 & 	87.8 & 	85.3 & 	82.5 \\
					\hline NN & 	98	 & 89.3 & 	88.2 & 	86.7 & 	85.1 \\
					\hline GCN & 98.3 & 	84.0 & 	83.7 & 	82.2 & 	78.8 \\
					\hline Proposed & \bf 	98.4 & \bf 	94.1	 & \bf 93.7 & \bf 	92.7 & \bf 	92.2 	\\
				\end{tabular} \\
				\hline
				PP & \begin{tabular}{@{}cccccc@{}}
					CNN& 97.5 & 	96.2 & 	96.1 & 	95.1 & 	94.6 \\
					\hline NN & 	95.6 & 	92.2 & 	90.3 & 	87.9 & 	85.9 \\
					\hline GCN & 		99.5 & 	96.5 & 	96.5 & 	96.6 & 	96.7 \\
					\hline Proposed & \bf 	99.9 & \bf 	99.6 & \bf 	99.4 & \bf 	99.2 & \bf 	98.4	\\
				\end{tabular} \\
				\hline
			\end{tabular}
		\end{table}
		\begin{table}[!ht]
			\centering
			\caption{LAR of SPG, DPG, PP  When Different States of Switches Change Network Topology}\label{topology}
			\begin{tabular}{c c c c c }
				\hline \hline SPG & \begin{tabular}{@{}cccc@{}}
					Switch & Close 7\&8 & Open 1-6 & Open 1-3 \\
					\hline 	CNN& 80.0	 & 84.4 & 	88.8 	\\
					\hline NN & 	75.0 & 	82.5	 & 81.7	 \\
					\hline GCN & 	56.9 & 	58.3 & 	59.6 \\
					\hline Proposed & \bf 	95.8 & \bf 	94.5 & \bf 	96.9 \\
				\end{tabular} \\
				\hline DPG & \begin{tabular}{@{}cccc@{}}
					Switch & Close 7\&8 & Open 1-6 & Open 1-3 \\
					\hline CNN& 84.7	 & 	88.3	 & 	90.3	 	\\
					\hline NN & 82.9	 & 	91.0		 & 89.3	 \\
					\hline GCN & 		80.5	 & 	66.9	 & 	85.6	 \\
					\hline Proposed & \bf 93.6	 & \bf 	94.4		 & \bf 96.5	 	\\
				\end{tabular} \\
				\hline
				PP & \begin{tabular}{@{}cccc@{}}
					Switch & Close 7\&8 & Open 1-6 & Open 1-3 \\
					\hline 	CNN& 96.1 & 	95.0 & 	96.9 	\\
					\hline NN & 		93.6 & 	94.1 & 	94.1 \\
					\hline GCN & 		91.4 & 	95.6 & 	97.3 \\
					\hline Proposed & \bf 	97.2 & \bf 	99.0	 & \bf 99.9 	\\
				\end{tabular} \\
				\hline
			\end{tabular}
		\end{table}
		\subsection{Extension to IEEE 37-node Test Feeder} \label{G}
		\begin{table}[!ht]
			\centering
			\caption{Location Performance of All Types of Faults When Using Various Training Strategies} \label{train}
			\begin{tabular}{c c c c c c }
				\hline \hline Supervised & \begin{tabular}{@{}ccccc@{}}
					$\beta$ & 75\% & 50\% & 25\% & 15\% \\
					\hline 	F1 Score & 91.0 & 	90.4 & 	88.1 & 	84.7 \\
					\hline LAR & 91.1 & 	90.3 & 	88.2 & 	84.8 \\
					\hline LAR$^{\text{1-hop}}$ & 97.6	 & 96.8	 & 96.5	 & 96.5 \\
				\end{tabular} \\
				\hline Proposed & \begin{tabular}{@{}ccccc@{}}
					$\beta$ & 75\% & 50\% & 25\% & 15\% \\ \hline
					F1 Score & \bf95.6& \bf	94.9& \bf	93.3& \bf	91.3 \\
					\hline LAR & \bf 95.6& \bf	94.9& \bf	93.3& \bf	91.3 \\
					\hline LAR$^{\text{1-hop}}$ & \bf98.9& \bf	98.5& \bf	99.1& \bf	98.9 \\
				\end{tabular} \\
				\hline
			\end{tabular}
		\end{table}

		We extend our graph framework to the IEEE 37-node test feeder \cite{K91}, 
		where 15 nodes are measured. To indicate the adaptation of our graph model, we keep the same graph structures described in Section \ref{structure}. Only the dimensions of inputs become $\bar{X}^p \in R^{36 \times 6}, p \in [1,\bar{N}], \bar{y}^p \in \{1, \cdots, \bar{c}\}, \bar{c} = 36$, where $\bar{X}^p$ only has 15 nonzero rows corresponding to those observed nodes. We generate $\bar{N} = 12960$ data samples in the 37-node test feeder, including SPG, DPG, and PP faults at all possible nodes, accompanied with load fluctuating. We train our graph framework with different percentages of datasets and test by the remaining data. The location performance of all types of faults is shown in Table~\ref{train} in the line of ``Proposed Training'', which denotes that we employ the proposed training strategy in Section \ref{T1_T2}. Comparably, ``Supervised Training'' denotes that the graph model is trained by the conventional supervise learning method, i.e., regard $\G_{\text{I}}$ as a whole and update all the trainable parameters in $\Theta_{\text{I}} $ in each iteration by optimizing \eqref{obj_I}.

		Table~\ref{train} shows that the proposed training strategy can enhance up to 6\% of LAR than that using ``Supervised Training''. Note that the similar effectiveness also appears in the 123-node test feeder. The intuition behind is that the ``local aggregation'' and ``global transformation'', functioning as the encoder and decoder of the graphical input data, have better convergence if trained alternatively \cite{GBC16, HYLJ17}.

\section{Conclusions and Future Works}\label{sec:con}
The black-box machine learning fails in power grids mainly due to the practical challenges: low observation, insufficient labeled datasets, and dynamic data distributions. This paper handles these issues by establishing a two-stage robust data-driven algorithm for fault location via embedding power grid physics into a graph learning framework.

We theoretically demonstrate the benefits of the proposed adjacency matrices to address the sparse observability and low label rates challenges. Experimental results illustrate the superior performances of the proposed approach over three baseline classifiers. A large number of OOD datasets validate our approach's robustness to load variations and topology changes.
Our future interest is to optimize the placement of PMUs to maximize location accuracy at the minimum cost.


\bibliographystyle{apalike}
\bibliography{sample}


\end{document}